\documentclass[conference]{IEEEtran}
\IEEEoverridecommandlockouts
\usepackage{cite}
\usepackage{amsmath,amssymb,amsfonts}
\usepackage{graphicx}
\usepackage{textcomp}
\usepackage{xcolor}
\def\BibTeX{{\rm B\kern-.05em{\sc i\kern-.025em b}\kern-.08em
    T\kern-.1667em\lower.7ex\hbox{E}\kern-.125emX}}

\usepackage{multirow, multicol}
 \usepackage{url}
\usepackage{algpseudocode}
\usepackage{algorithm}
\usepackage{pbox}
\usepackage{stfloats}
\usepackage{subcaption}
\usepackage{amsmath}
\usepackage{hyperref}

\begin{document}

\title{Enhancing Satellite Object Localization with Dilated Convolutions and Attention-aided Spatial Pooling\\
% {\footnotesize \textsuperscript{*}Note: Sub-titles are not captured in Xplore and should not be used}
% \thanks{Identify applicable funding agency here. If none, delete this.}
\thanks{This work is supported by NSF grant OAC–1942714, and NASA grants 80NSSC22K0641 and 80NSSC21M0027. We would like to thank Dr. Jinbo Wang and Benjamin Holt from NASA JPL for providing us with the Ocean Eddy data.}
}

\author{\IEEEauthorblockN{Seraj Al Mahmud Mostafa$^{1}$, Chenxi Wang$^{1}$, Jia Yue$^{2,3}$, Yuta Hozumi$^3$, Jianwu Wang$^{1}$}
\IEEEauthorblockA{\textit{$^1$Department of Information Systems, University of
Maryland, Baltimore County,} Baltimore, MD, USA\\
\textit{$^2$Department of Physics, Catholic University of America,} Washington DC, USA\\
\textit{$^3$NASA Goddard Space Flight Center,} Greenbelt, MD, USA\\
$^{1}$\{serajmostafa, chenxi, jianwu\}@umbc.edu, $^2$\{jia.yue, yuta.hozumi\}@nasa.gov}
}

\maketitle

\begin{abstract}
Object localization in satellite imagery is particularly challenging due to the high variability of objects, low spatial resolution, and interference from noise and dominant features such as clouds and city lights. In this research, we focus on three satellite datasets: upper atmospheric Gravity Waves (GW), mesospheric Bores (Bore), and Ocean Eddies (OE), each presenting its own unique challenges. These challenges include the variability in the scale and appearance of the main object patterns, where the size, shape, and feature extent of objects of interest can differ significantly.
To address these challenges, we introduce YOLO-DCAP, a novel enhanced version of YOLOv5 designed to improve object localization in these complex scenarios. YOLO-DCAP incorporates a Multi-scale Dilated Residual Convolution (MDRC) block to capture multi-scale features at scale with varying dilation rates, and an Attention-aided Spatial Pooling (AaSP) module to focus on the global relevant spatial regions, enhancing feature selection. 
These structural improvements help to better localize objects in satellite imagery. Experimental results demonstrate that YOLO-DCAP significantly outperforms both the YOLO base model and state-of-the-art approaches, achieving an average improvement of 20.95\% in mAP50 and 32.23\% in IoU over the base model, and 7.35\% and 9.84\% respectively over state-of-the-art alternatives, consistently across all three satellite datasets. These consistent gains across all three satellite datasets highlight the robustness and generalizability of the proposed approach.
% These structural improvements help to better localize objects in satellite imagery. Experimental results demonstrate that YOLO-DCAP significantly outperforms state-of-the-art models under challenging conditions across all three satellite datasets, highlighting the generalizability of the approach. 
Our code is open sourced at \href{https://github.com/AI-4-atmosphere-remote-sensing/satellite-object-localization}{https://github.com/AI-4-atmosphere-remote-sensing/satellite-object-localization}.
\end{abstract}
\vspace{.5em}


\section{Introduction}
Satellite data plays a vital role in Earth informatics, offering valuable insights into the planet's environment and climate. Specifically, satellite imagery has garnered significant attention for its ability to help scientists interpret a range of phenomena, including atmospheric conditions, sea surface changes, and environmental shifts. Gravity Waves (GW), Bore in the atmosphere and oceanic events such as Ocean Eddies are crucial in monitoring these factors in the Earth's surface and atmosphere. Upper atmospheric gravity waves (GW), distinct from gravitational waves \cite{jovanovic2018nature} are oscillations caused by buoyancy or gravity's restoring force, generating physical disturbances in the atmosphere. Disturbances like airflow over mountains, jet streams, and thunderstorms generate atmospheric GWs, which displace air parcels and produce wave patterns akin to ripples on water \cite{mann2019}.
Mesospheric Bores (Bore) \cite{hozumi2019geographical}, are a specific type of wave characterized by sharp discontinuities in airglow brightness with trailing wave structures. These phenomena are created by large-amplitude GWs propagating within mesospheric wave ducts, providing crucial insights into wave-duct interactions and energy transfer in the upper atmosphere. Both the GW and Bore significantly impact the middle atmosphere, influencing turbulence and tidal waves \cite{Fritts2003}. The third focus is on small-scale OEs \cite{ivanov2002oceanic}, 
% \cite{mostafa2023cnn}, 
which are circular water currents ranging from a few kilometers to over 300 kilometers. These mesoscale eddies are vital for horizontal water transport, heat, and tracers, while smaller-scale eddies (below 50 km) are crucial for vertical mixing and climate system interactions \cite{CHELTON2011167}. 
% Detailed information about these phenomena are in the corresponding references.

Despite the wealth of information these satellite datasets provide, they present significant complexities \cite{guo2007semantic}. The satellite data poses substantial difficulties, particularly in the GW and Bore datasets, which are affected by various interferences such as city lights, clouds, and instrumental noise, which obscure subtle atmospheric phenomena \cite{mostafa2025gwavenet, gonzalez2022atmospheric}. In contrast, the OE dataset is primarily influenced by generic noise and instrumental artifacts \cite{mostafa2023cnn}. These datasets present two fundamental challenges: \textbf{\textit{(1)}} significant variability in the scale, shape, and extent of the main object patterns, and, \textbf{\textit{(2})} the objects of interest such as GW, Bore, or OE can be mixed with or hindered by interference, including occlusion and overlap interference, complicating the detection process. Additionally, a common issue across all of these datasets is they are captured and stored within a single band, making it challenging to distinguish and remove unwanted objects.

To address these challenges, this study proposes YOLO-DCAP (\underline{D}ilated \underline{C}onvolution and \underline{A}ttention-aided \underline{P}ooling), an enhanced version of YOLOv5 \cite{yolov5v6}. YOLO-DCAP integrates two advanced techniques:
\textit{\textbf{(1)} \underline{M}ulti-scale \underline{D}ilated \underline{R}esidual \underline{C}onvolution (MDRC)}, and, \textit{\textbf{(2)} \underline{A}ttention-\underline{a}ided \underline{S}patial \underline{P}ooling (AaSP)}. MDRC captures features across multiple scales using dilated convolutions, expanding the receptive field without reducing resolution. The inclusion of residual connections preserves and propagates features across layers, enabling more robust feature extraction at scale. This approach is particularly effective for identifying complex structures across various spatial resolutions in all the datasets used in this study. AaSP on the other hand, enhances feature learning by focusing on the most relevant spatial regions. The attention mechanism emphasizes critical features while suppressing irrelevant ones, refining the spatial adaptive pooling process. This allows the model to concentrate on essential areas, including regions affected by occlusion or overlap interference, improving feature selection, spatial awareness, and ensuring the retention of crucial details during pooling.

The paper is organized as follows. Section \ref{sec:background} introduces the datasets and base YOLO model, Section \ref{sec:rw} reviews related studies, Section \ref{sec:methods} details the methodology, and Section \ref{sec:exp} presents the experiments. Finally, Section \ref{sec:disc-and-conc} concludes our work.

\begin{figure*}[htbp]
    \centering
    \begin{tabular}{c@{\hspace{2mm}}c@{\hspace{2mm}}c@{\hspace{2mm}}c@{\hspace{2mm}}c@{\hspace{2mm}}c@{\hspace{2mm}}c@{\hspace{2mm}}c}
        & \newcommand{\smaller}\textbf{Labeled Data} 
        & \newcommand{\smaller}\textbf{YOLO-DCAP} & \newcommand{\smaller}\textbf{SPP \cite{he2015spatial}} & \newcommand{\smaller}\textbf{CBAM \cite{woo2018cbam}} & \newcommand{\smaller}\textbf{ViT \cite{dosovitskiy2020image}} & \newcommand{\smaller}\textbf{Transformer \cite{vaswani2017attention}} & \newcommand{\smaller}\textbf{YOLO \cite{yolov5v6}} \\ %[2ex]
        % {\rotatebox{90}{\textbf{Gravity Wave}}} &
        \multirow{1}{*}[2cm]{\rotatebox{90}{\textbf{Gravity Wave}}} & 
        \includegraphics[width=0.125\linewidth, height=2.5cm]{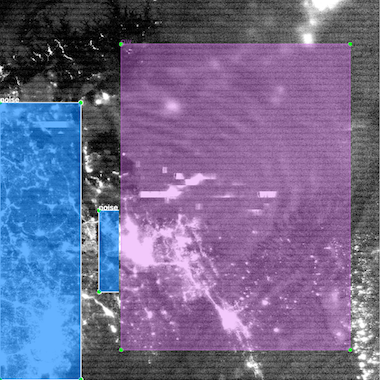} &
        \includegraphics[width=0.125\linewidth, height=2.5cm]{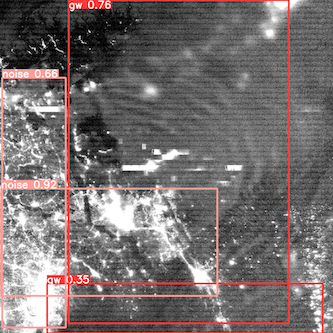} &
        \includegraphics[width=0.125\linewidth, height=2.5cm]{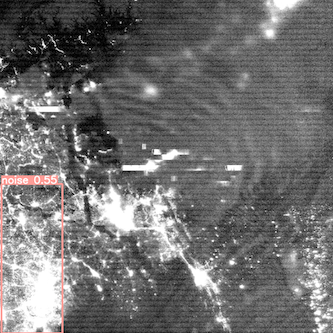} &
        \includegraphics[width=0.125\linewidth, height=2.5cm]{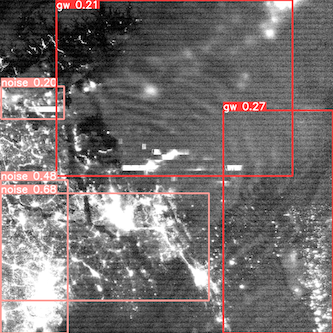} &
        \includegraphics[width=0.125\linewidth, height=2.5cm]{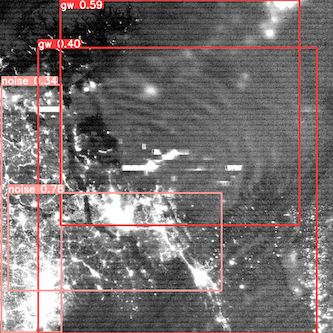} &
        \includegraphics[width=0.125\linewidth, height=2.5cm]{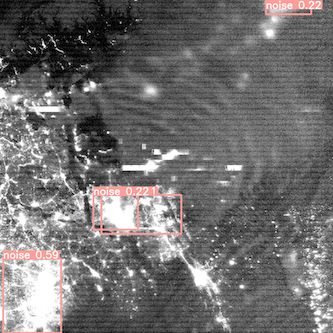} &
        \includegraphics[width=0.125\linewidth, height=2.5cm]{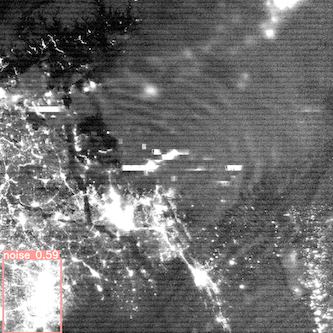} \\ %[2ex]
        % \rotatebox{90}{\textbf{Bore}} &
        \multirow{1}{*}[1.2cm]{\rotatebox{90}{\textbf{Bore}}} & 
        \includegraphics[width=0.125\linewidth, height=2.5cm]{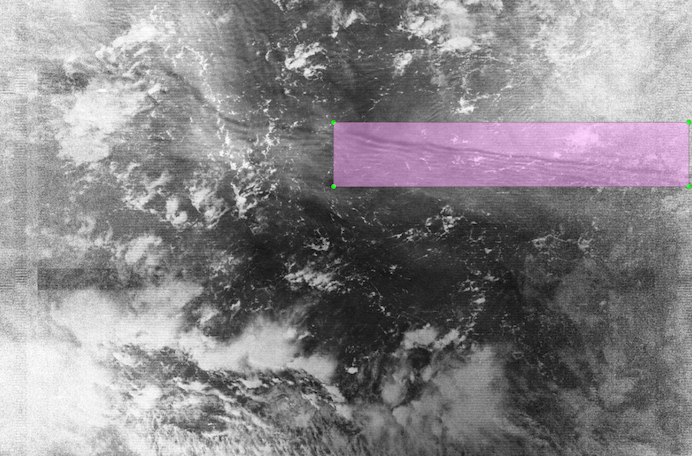} &
        \includegraphics[width=0.125\linewidth, height=2.5cm]{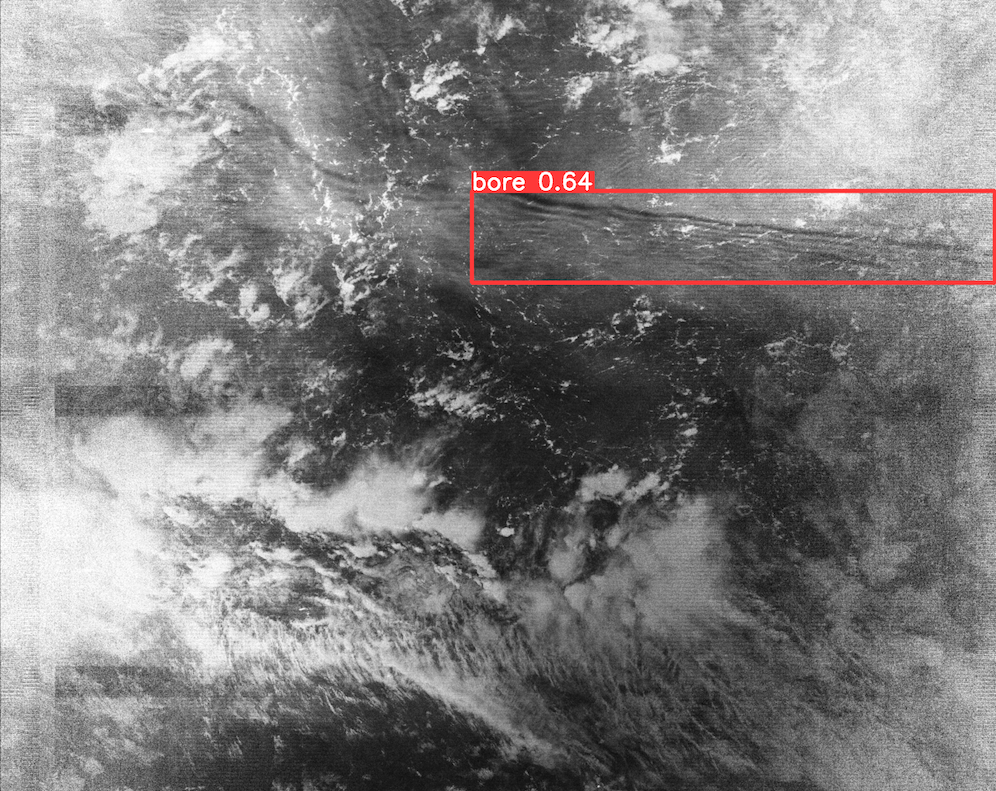} &
        \includegraphics[width=0.125\linewidth, height=2.5cm]{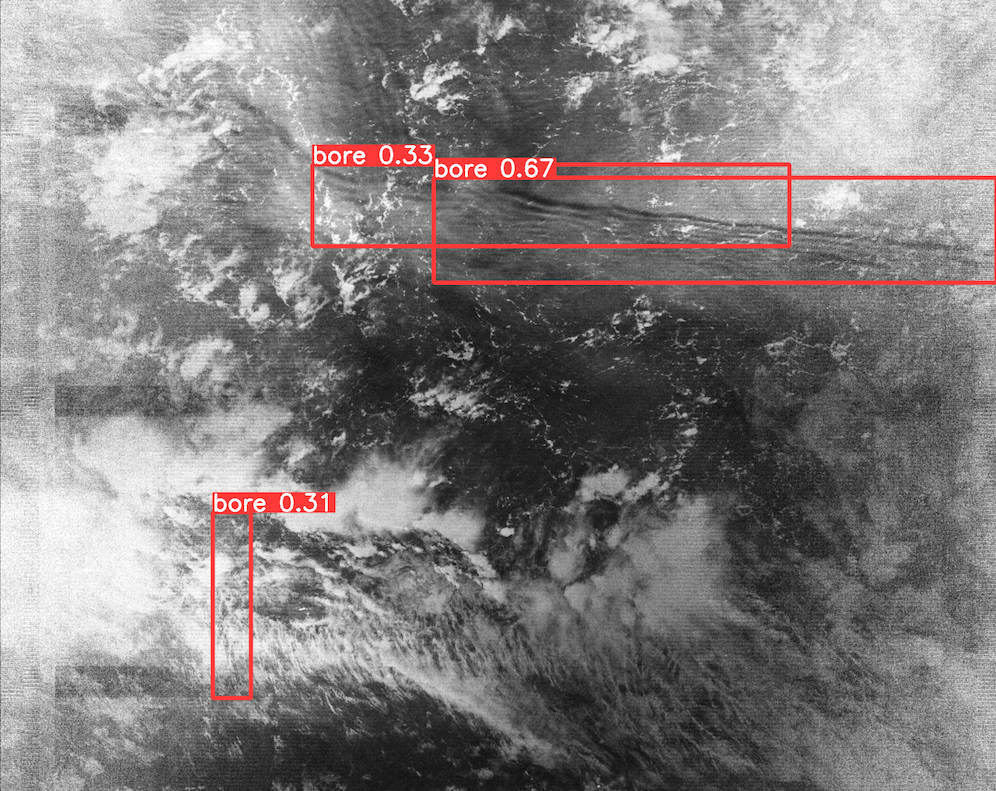} &
        \includegraphics[width=0.125\linewidth, height=2.5cm]{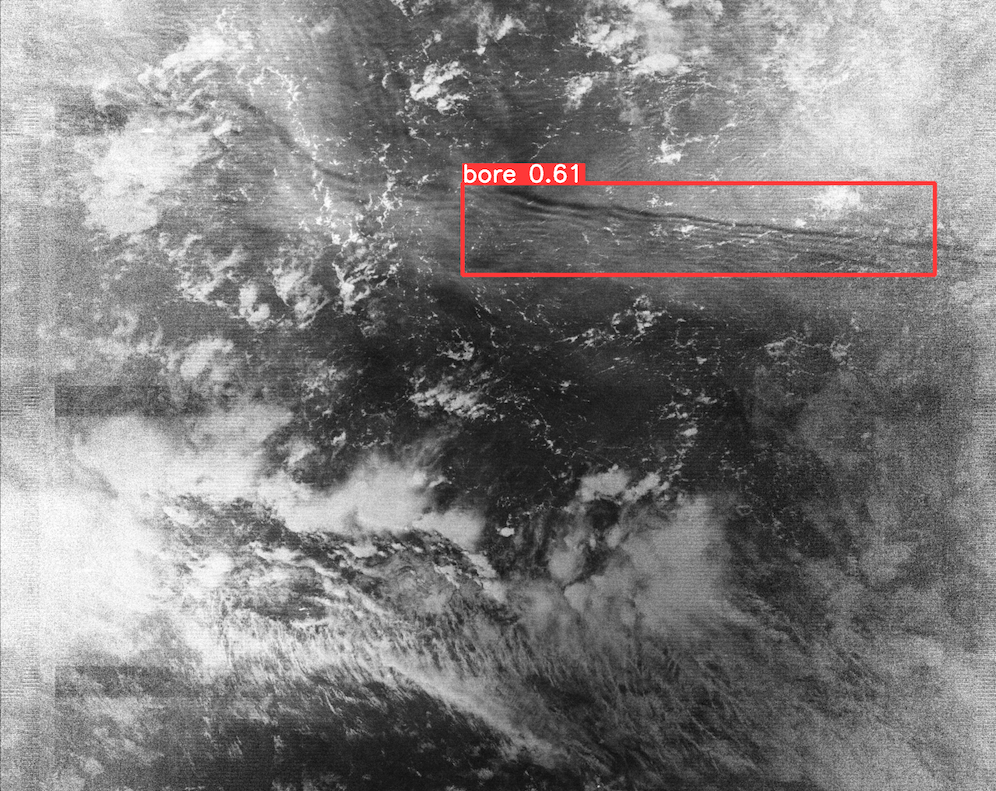} &
        \includegraphics[width=0.125\linewidth, height=2.5cm]{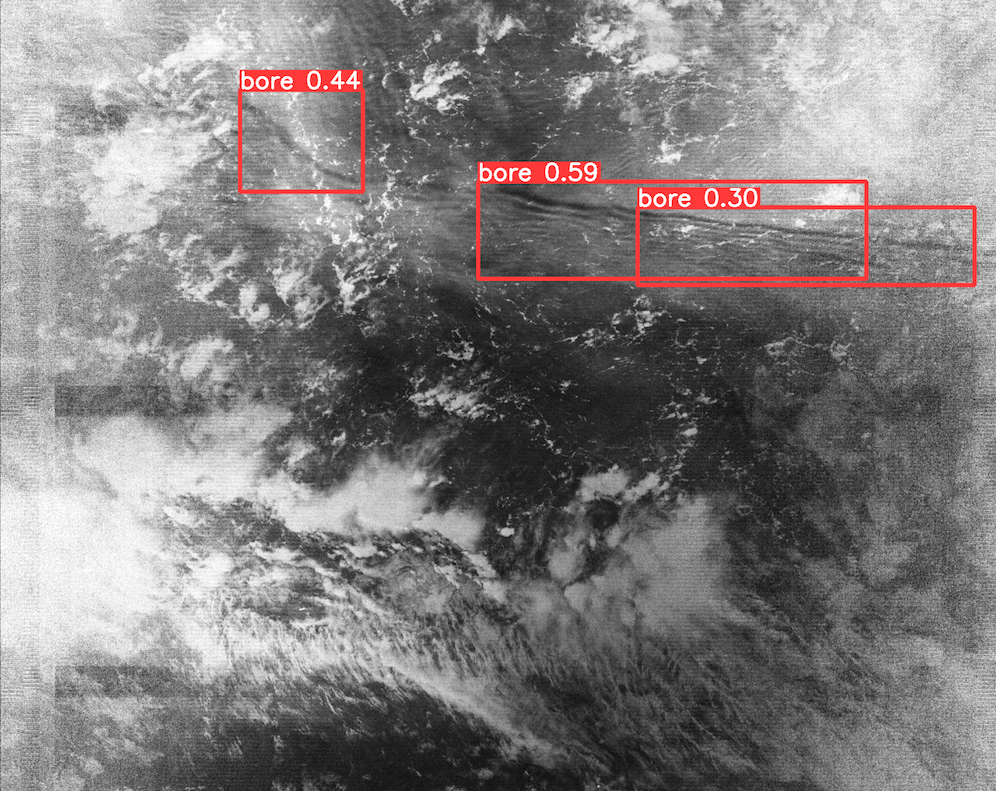} &
        \includegraphics[width=0.125\linewidth, height=2.5cm]{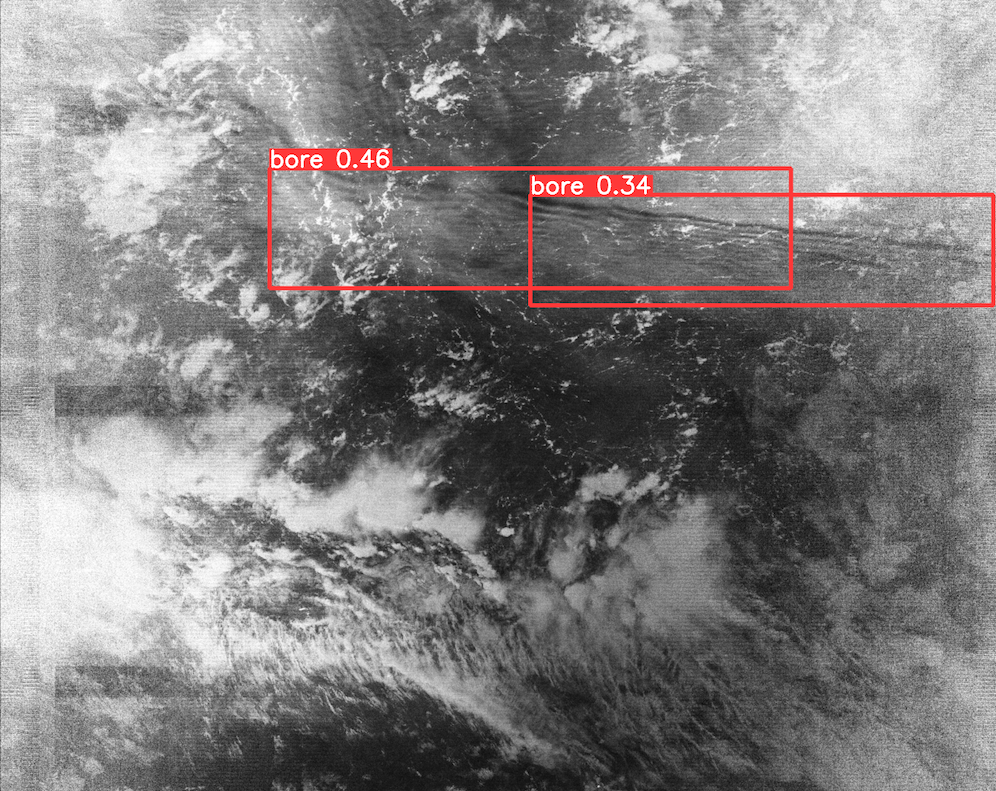} &
        \includegraphics[width=0.125\linewidth, height=2.5cm]{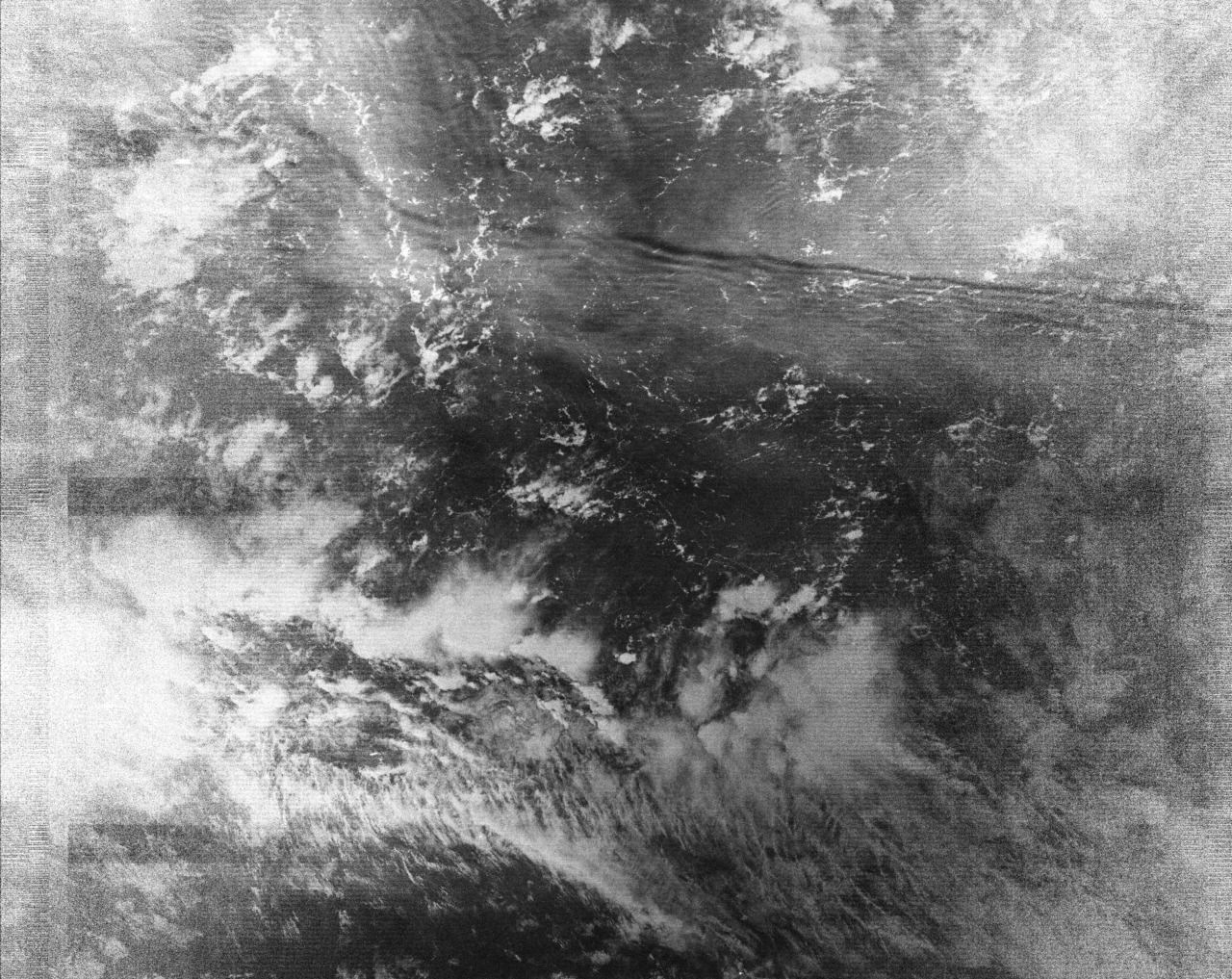} \\%[2ex]
        % \rotatebox{90}{\textbf{Ocean Eddy}} &
        \multirow{1}{*}[1.9cm]{\rotatebox{90}{\textbf{Ocean Eddy}}} &
        \includegraphics[width=0.125\linewidth, height=2.5cm]{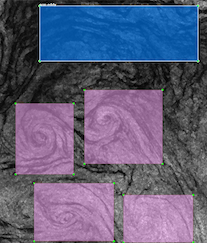} &
        \includegraphics[width=0.125\linewidth, height=2.5cm]{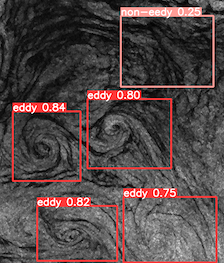} &
        \includegraphics[width=0.125\linewidth, height=2.5cm]{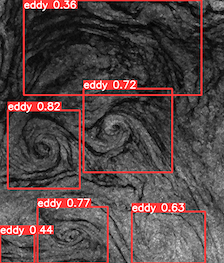} &
        \includegraphics[width=0.125\linewidth, height=2.5cm]{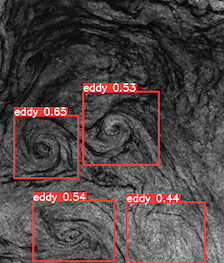} &
        \includegraphics[width=0.125\linewidth, height=2.5cm]{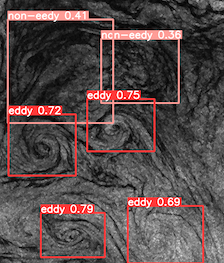} &
        \includegraphics[width=0.125\linewidth, height=2.5cm]{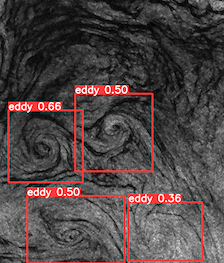} &
        \includegraphics[width=0.125\linewidth, height=2.5cm]{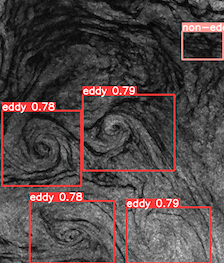}
    \end{tabular}
    \caption{Localization comparison across baseline models and proposed YOLO-DCAP on GW, Bore, and OE datasets. In the `Labeled Data', pink regions represent objects of interest (GW, Bore, or OE), while blue regions indicate non-objects of interest such as city lights and clouds (labeled as `noise' in GW and Bore data, and as `non-eddy' in OE data).}
    \label{fig:vis-res}
\end{figure*}

\section{Background}
\label{sec:background}

\subsection{Datasets}
The datasets used in this study underwent specific preprocessing steps tailored to their unique characteristics. The GW and Bore datasets are based on the night band data from the Visible Infrared Imaging Radiometer Suite (VIIRS) Day/Night Band (DNB) aboard the Suomi NPP satellite.
% \cite{gravity_wave_data}. 
OE dataset on the other hand is the  Synthetic Aperture Radar (SAR) satellite images.
GW data captured at extremely low radiance levels $10^{-9}$ W/$cm^{-2}$ $sr^{-1}$ \cite{murphy2006visible}.
% , gravitywavedata}. 
These datasets, originally stored in HDF5 format
% \cite{hdf5}, 
were processed through several steps including minimum value subtraction, median scaling, and normalization to 0.5, while transforming the intensity distribution from normal to uniform \cite{mostafa2025gwavenet}. 
Bore data processing started with level-1B DNB data, converting 4-byte float data to 1-byte integer (0-255) format, followed by resizing to 1,280 x 1,017 pixels (approximately 2.5 km resolution) using bilinear interpolation \cite{hozumi2019geographical}. The Bore dataset, comprising 916 DNB images with 988 frontal wave objects from January 2013 to July 2023, was specifically collected during moon-free nights to avoid interference with airglow detection. 
OE data are Synthetic Aperture Radar (SAR) satellite images, originally in TIFF format, were processed using Quantum Geographic Information System (QGIS) software and the Geospatial Data Abstraction Library (GDAL). These tools facilitated the conversion of TIFF images to a more manageable unsigned integer (UInt8) PNG format, significantly enhancing our ability to analyze detailed SAR data efficiently \cite{mostafa2023cnn}.
All datasets were ultimately standardized to PNG format to facilitate uniform processing while preserving their essential characteristics.

\subsection{YOLO (You Only Look Once)}
We employ YOLOv5 \cite{yolov5v6}, a cutting-edge object detection model that combines feature extraction, localization, and classification in a single framework. As one of the prominent DNN-based object detection models, YOLO excels at capturing features within a given set of regions \cite{gabale2018cdssd}. Its backbone is designed using the CSPDarknet53 that uses convolution, C3 (cross convolution), and Spatial Pyramid Pooling-Faster (SPPF), a faster variant of SPP, is mathematically identical but requires fewer FLOPs for feature extraction. The neck, featuring SPP and CSP-PAN layers, enhances these features further, and the detection head outputs bounding boxes, objectness scores, and class probabilities.  The model enhances its efficacy through strategies like multi-scale training, automatic anchor adjustments, cosine learning rate adjustments, and mixed-precision training. Its loss function integrates binary cross-entropy for objectness and classification with CIoU loss for precise localization.

\section{Related Works}
\label{sec:rw}

\subsection{Satellite Events}
Gravity wave studies focusing on their observation and characteristics are discussed in \cite{miller2015upper, coisson2015first}. Advanced approaches using deep learning for detection and classification with improved accuracy are presented in \cite{gonzalez2022atmospheric, mostafa2025gwavenet}. Similarly, bore events have been observed and characterized by Fechine et al. \cite{fechine2005mesospheric, hozumi2019geographical}, who analyzed the associated mesospheric dynamics, while Smith et al. \cite{smith2017characterization} examined their structure and behavior, collectively advancing our understanding of these phenomena. Ocean eddies have been extensively studied, with detection and segmentation tasks performed using deep convolutional networks \cite{santana2022oceanic, moschos2023computer}, and advanced techniques such as YOLO models employed for eddy localization \cite{zi2024ocean, mostafa2024yolo}.

\subsection{Dilated Convolutions}
Dilated convolutions have demonstrated effectiveness across various applications. Zhou et al. showed that dilated convolutions expand the receptive field of feature points without compromising feature map resolution \cite{zhou2018d}. Liu et al. improved road area extraction in semantic segmentation by combining dilated convolutions with residual learning \cite{liu2019d}. Zhang et al. utilized dilated convolutions in atrous CNNs to capture more semantic information for ultrasound image segmentation \cite{zhang2020multiple}. Chen et al. investigated optimal dilation rates for aggregating multiscale features and enlarging the receptive field \cite{chen2021effective}. In \cite{chen2017deeplab}, the authors demonstrated that combining dilated convolutions with spatial pyramid pooling preserves feature resolution while maintaining a large receptive field, effectively segmenting objects at multiple scales. Li et al. introduced self-smoothing atrous convolution to enhance the receptive field \cite{li2021cascaded}, while Wang et al. developed smoothing techniques to address gridding artifacts in dense predictions \cite{wang2018smoothed}.

\subsection{Attention Mechanisms}
Attention mechanisms have enhanced object detection and semantic segmentation performance. Park et al. developed a pyramid attention mechanism to improve detection across feature pyramid networks \cite{park2022pyramid}. Zhou et al. proposed the Scale-aware Spatial Pyramid Pooling (SSPP) module, alongside Encoder Mask and Scale-Attention modules, addressing challenges in scale-awareness, boundary sharpness, and long-range dependency modeling \cite{zhou2020scale}. Cao et al. designed a network incorporating a Context Extraction Module and an Attention-guided Module to enhance object localization and recognition by leveraging contextual information and adaptive attention mechanisms \cite{cao2020attention}.

\subsection{Synergizing Dilated Convolutions, Attention Mechanisms, and Spatial Pyramid Pooling}
Integration of dilated convolutions, attention mechanisms, and spatial pyramid pooling has been explored to improve feature extraction and spatial information pooling. Feng et al. introduced AttSPP-net, combining a soft attention mechanism with Spatial Pyramid Pooling (SPP) for action recognition, allowing the network to focus on relevant regions and improving robustness to action deformation \cite{feng2017attention}. Guo et al. developed the Spatial Pyramid Attention Network (SPANet), which incorporates a spatial pyramid structure in the attention path, enhancing CNN performance for image recognition \cite{guo2020spanet}. Qui et al. created the Attentive Atrous Spatial Pyramid Pooling (A2SPP) method, combining Channel-Embedding Spatial Attention (CESA) and Spatial-Embedding Channel Attention (SECA) to adapt to different feature scales \cite{qiu2022a2sppnet}. Similarly, Wang et al. developed a salient object detection network using multi-scale saliency attention, focusing on salient regions and achieving state-of-the-art performance with rapid inference speed \cite{wang2019salient}.

The proposed YOLO-DCAP framework distinctly differs from the methods discussed above, presents a novel approach to enhancing object localization in satellite imagery. It tackles challenges such as diverse object shapes, sizes, patterns, occlusion, and overlap through the integration of Multi-scale Dilated Residual Convolution (MDRC) and Attention aided Attention Pooling (AaSP). MDRC captures detailed features across multiple scales, while AaSP maintains a global context for robust performance in complex scenarios. In addition, our advanced detection mechanism localizes satellite objects with higher precision, significantly outperforming state-of-the-art methods in real-world applications.

\section{Methodology}
\label{sec:methods}
% \vspace{-.5em}

\subsection{YOLO-DCAP}
\begin{figure*}[ht]
  \centering
  
  % Left column: overall architecture (a)
  \begin{minipage}[c]{0.42\textwidth}
    \centering
    \begin{subfigure}{\textwidth}
      \centering
      \includegraphics[height=0.42\textheight]{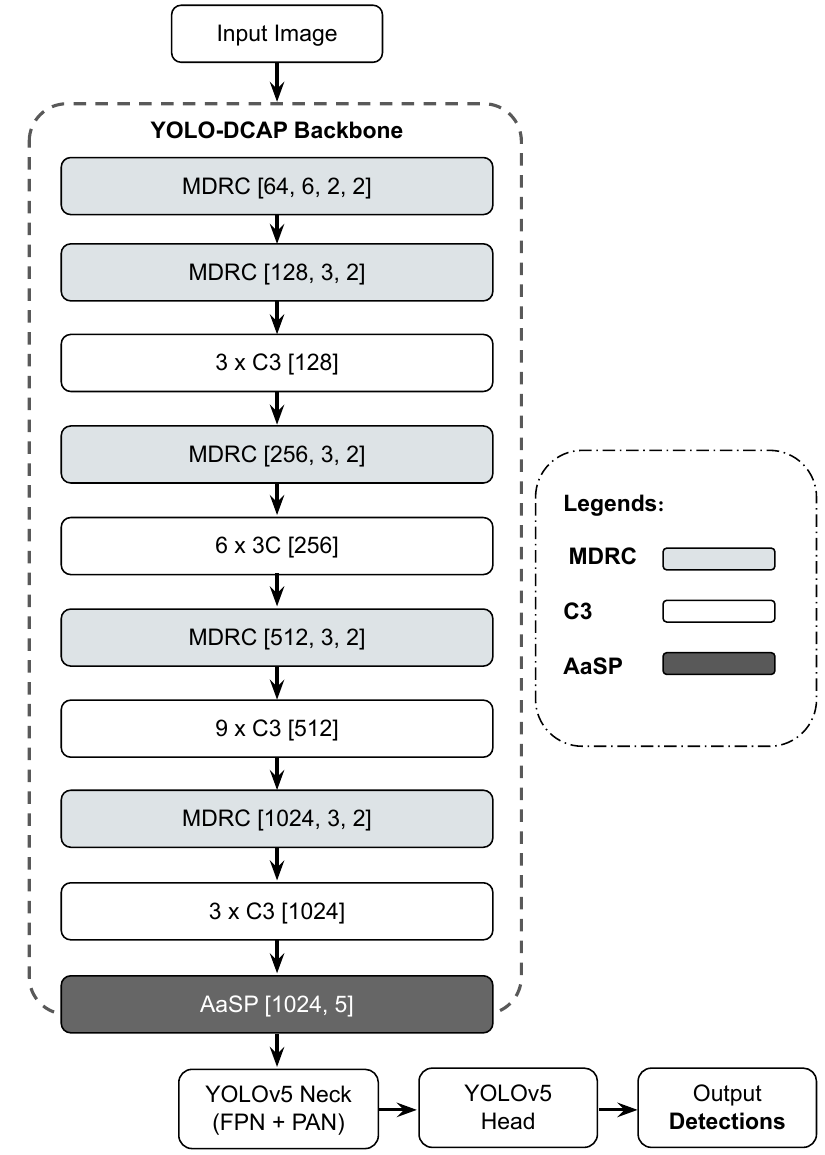}
      \caption{YOLO-DCAP backbone with MDRC and AaSP}
      \label{subfig:yolo-a}
    \end{subfigure}
  \end{minipage}
  \hfill
  % \hspace{-.1em}
  % Right column: top (b) and bottom (c)
  \begin{minipage}[c]{0.57\textwidth}
    % \vspace{0.1\textheight}  % Negative value to move up
    \centering
    % Top image: MDRC (b)
    \begin{subfigure}{\textwidth}
      \centering
      \includegraphics[height=0.1\textheight]{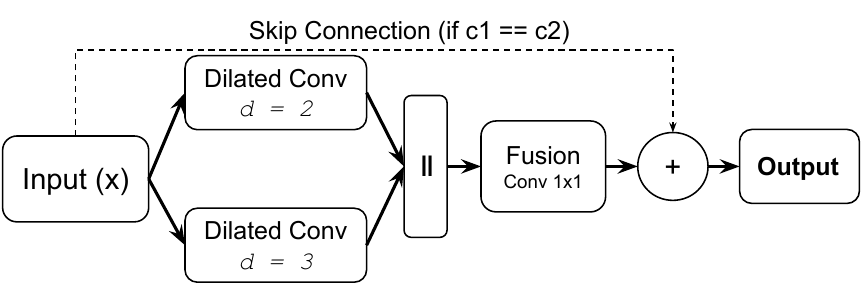}
      \caption{Multi-scale Dilated Residual Convolution (MDRC) block}
      \label{subfig:mdrc-b}
    \end{subfigure}
    
    \vspace{0.005\textheight}
    
    % Bottom image: SSCA (c)
    \begin{subfigure}{\textwidth}
      \centering
      \includegraphics[height=0.31\textheight]{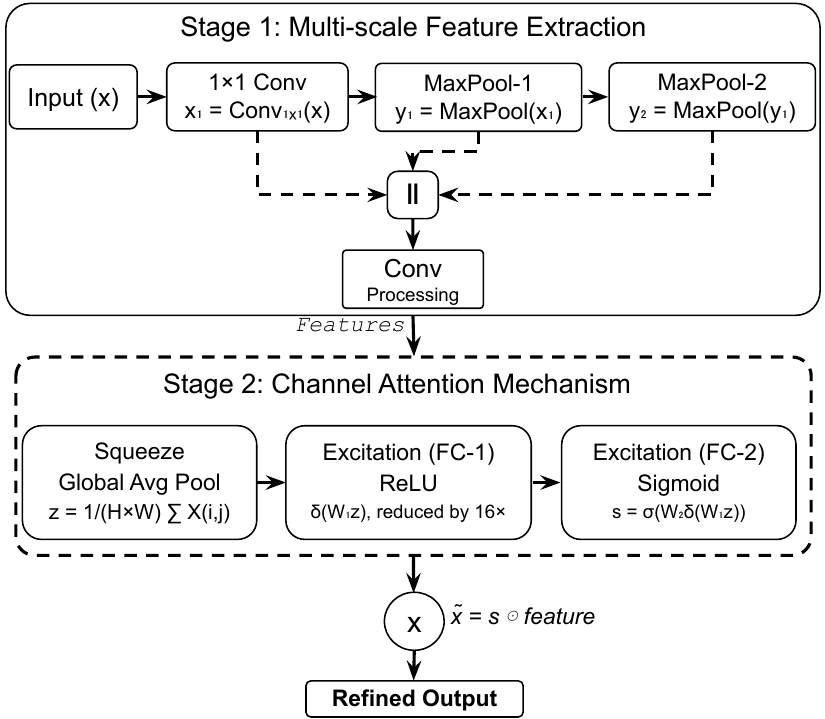}
      \caption{Attention-aided Spatial Pooling (AaSP)}
      \label{subfig:ssca-c}
    \end{subfigure}
  \end{minipage}
  
  \caption{Proposed YOLO-DCAP architecture for satellite object detection in satellite imagery: (a) YOLO-DCAP backbone with MDRC and AaSP, (b) MDRC for capturing multi-scale features, and (c) AaSP to enhance global feature pooling by focusing on relevant spatial regions. `$||$' denotes concatenation.}
  \label{fig:aasp}
\end{figure*}

The proposed YOLO-DCAP architecture, shown in Figure \ref{fig:aasp}, integrates the modified YOLO backbone, Multi-scale Dilated Residual Convolution (MDRC), and the Attention-aided Spatial Pooling (AaSP) module, illustrated in Figure~\ref{subfig:yolo-a}, Figure~\ref{subfig:mdrc-b}, and Figure~\ref{subfig:ssca-c}, respectively. In our design, MDRC blocks replace all conventional convolutional layers in the YOLOv5 backbone. Unlike standard convolutions with fixed receptive fields, MDRC leverages varying dilation rates to capture objects at multiple scales without increasing computational complexity. Additionally, the Spatial Pyramid Pooling layer is replaced with the AaSP module to further strengthen feature extraction and emphasize global object context through attention mechanisms.

\subsection{Multi-scale Dilated Residual Convolution (MDRC)}

In our approach, we replace all conventional convolutional layers in the YOLOv5 backbone with our novel Multi-scale Dilated Residual Convolution (MDRC) block. Traditional convolutions are limited by a fixed receptive field, which can impede the detection of objects with variable scales and complex patterns. The proposed MDRC addresses this limitation by employing parallel dilated convolutions with dilation rates, typically set to \([2, 3]\), to capture multi-scale features.
For each branch, the dilated convolution operation is defined as:
\begin{equation}
y[i] = \sum_{k} w[k] \cdot x[i + d \cdot k],
\end{equation}
\noindent where \(d\) denotes the dilation rate, \(w\) represents the convolution kernel, and \(x\) is the input.

The outputs from these parallel convolutions are concatenated along the channel dimension and then fused using a convolution (\(1\times1\)) to reduce the channel dimension back to \(c_2\), expressed as: \(\text{out} = \text{Conv}_{1\times1}\left(\text{Concat}\left(\{\text{Conv}_d(x) \mid d \in \text{dilations}\}\right)\right).\)
A key feature of MDRC is its residual connection, which is activated when the input and output channel dimensions are equal (\(c_1 = c_2\)). This is mathematically represented as: \( y = x + F(x, \{W_i\}) \), where \( F(x, \{W_i\}) \), where \(F(x, \{W_i\})\) denotes the concatenated outputs from the dilated convolutions with their respective weights \(\{W_i\}\). This residual connection preserves essential information and facilitates improved gradient flow during training.

\subsection{Attention-aided Spatial Pooling (AaSP)} 
The Attention-aided Spatial Pooling (AaSP) module enhances the Spatial Pyramid Pooling (SPP) framework \cite{he2015spatial} by introducing architectural modifications and incorporating an attention mechanism inspired by Squeeze-and-Excitation (SE) \cite{vosco2021tiled}. This enables the network to focus on the most informative features and adapt to varying input characteristics.

%%%%%%%%% Old description %%%%%%%%%%%%%%%
% The AaSP module has two stages, \textit{multi-scale feature extraction} and \textit{attention-based refinement}. In the first stage, the input feature map \( x \) is reduced along the channel dimension using a \( 1 \times 1 \) convolution: \( x_1 = \text{Conv}_{1 \times 1}(x) \). 
% % \begin{equation}
% % x_1 = \text{Conv}_{1 \times 1}(x)
% % \end{equation}
% Unlike traditional SPP, this reduces computational complexity. Next, two sequential max-pooling layers capture multi-scale features: \( y_i = \text{MaxPool}_{k \times k}(y_{i-1}), \quad i \in \{1, 2\} \), 
% % \begin{equation}
% % y_i = \text{MaxPool}_{k \times k}(y_{i-1}), i \in \{1, 2\}
% % \end{equation}
% where \( y_0 = x_1 \) and \( k \) is the kernel size. The first pooling focuses on fine-grained features, and the second captures global context. Sequential pooling replaces SPP's parallel pooling, prioritizing global features with reduced computation. The resulting features are processed along the channel dimension via an additional convolution.

The AaSP module has two stages: \textit{multi-scale feature extraction} and \textit{attention-based refinement}. First, the input feature map \( x \) is reduced via \( 1 \times 1 \) convolution: \( x_1 = \text{Conv}_{1 \times 1}(x) \), reducing computational complexity compared to traditional SPP having multiple parallel pooling. Two sequential max-pooling layers then capture multi-scale features: \( y_i = \text{MaxPool}_{k \times k}(y_{i-1}), i \in \{1, 2\} \), where \( y_0 = x_1 \) and pooling kernels increase in size. The first pooling extracts fine-grained features while the second captures global context. This sequential approach prioritizes global features with reduced computation. Multi-scale features \( \{x_1, y_1, y_2\} \) are concatenated and processed via convolution for the final output.

The second stage applies an attention mechanism for adaptive feature refinement \cite{hu2018squeeze} to the output features
\( X \) from the first stage. In the \textit{squeeze} phase, global spatial information is condensed into a channel descriptor through global average pooling:
\begin{equation}
z = \frac{1}{H \times W} \sum_{i=1}^{H} \sum_{j=1}^{W} X(i,j),
\end{equation}
\noindent where \( H \) and \( W \) are the spatial dimensions of \( X \). In the \textit{excitation} phase, channel interdependencies are captured using two fully connected layers with ReLU and sigmoid activations: \( s = \sigma(W_2 \delta(W_1 z)) \),
where \( \delta \) is ReLU, \( \sigma \) is sigmoid, and \( W_1, W_2 \) are learnable weights. The first layer reduces channel dimension by a factor of 16, while the second restores it to produce channel-specific importance weights.
Finally, in the \textit{scale} phase, these weights recalibrate feature maps via element-wise multiplication: \( \tilde{x} = s \odot X \),
where \( \tilde{x} \) is the refined output emphasizing important channels while suppressing less relevant ones. This selective feature enhancement improves model representational power by dynamically focusing attention on the most informative features.

% The second stage applies an attention mechanism for adaptive feature refinement \cite{hu2018squeeze}. In the \textit{squeeze} phase, global spatial information is condensed into a channel descriptor through global average pooling:
% \begin{equation}
% z = \frac{1}{H \times W} \sum_{i=1}^{H} \sum_{j=1}^{W} X(i,j),
% \end{equation}
% \noindent where \( H \) and \( W \) are the spatial dimensions of \( X \). In the \textit{excitation} phase, channel interdependencies are captured using two fully connected layers with ReLU and sigmoid activations: \( s = \sigma(W_2 \delta(W_1 z)) \),
% where \( \delta \) is ReLU, \( \sigma \) is sigmoid, and \( W_1, W_2 \) are learnable weights. The first layer reduces channel dimension by a factor of 16, while the second restores it to produce channel-specific importance weights.
% Finally, in the \textit{scale} phase, these weights recalibrate feature maps via element-wise multiplication: \( \tilde{x} = s \odot \text{out} \),
% where \( \tilde{x} \) is the refined output emphasizing important channels while suppressing less relevant ones. This selective feature enhancement improves model representational power by dynamically focusing attention on the most informative features.

\begin{table*}[hb]
  \centering 
  % \large
  % \scriptsize
  % \footnotesize
  \caption{Comparison of the state-of-the-art baseline methods with proposed MDRC, AaSP and YOLO-DCAP (combining both MDRC and AaSP) enhanced models across GW, Bore, and OE datasets.}
  \label{tab:sota-comp}
  \begin{tabular}{|c|l|c|c|c|c|c|} % Left and right borders
    \hline
    \textbf{Datasets} & \multicolumn{1}{c|}{\textbf{Methods}} & \textbf{Precision} & \textbf{Recall} & \textbf{mAP50 (\%)} & \textbf{mAP50-95 (\%)} & \textbf{IoU(\%)}\\
    % Dataset & Method & Precision (\%) & Recall (\%) & mAP50 (\%) & mAP50-95 (\%) & IoU (\%)\\ 
    \hline
    \multirow{12}{*}{GRAVITY WAVE} %{\rotatebox{90}{GRAVITY WAVE}} 
    & Y\textsubscript{base} & 56.30 & 41.30 & 41.80 & 16.30 & 32.76\\
    \cline{2-7}
    & Y\textsubscript{base}$$+$$MDRC (ours) & 56.20 & 39.70 & 47.80 & 16.30 & 40.62\\
    \cline{2-7}
    & Y\textsubscript{base}$+$Transformer & 47.40 & 48.70 & 44.50 & 11.40 & 44.22 \\
    & Y\textsubscript{base}$+$CBAM & 39.30 & 52.40 & 41.60 & 14.30 & 42.10 \\
    & Y\textsubscript{base}$+$ViT & 44.10 & 55.20 & 44.00 & 14.90 & 42.76\\
    & Y\textsubscript{base}$+$SPP & 59.10 & 57.10 & 53.60 & 21.30 & 51.28\\
    & Y\textsubscript{base}$+$AaSP (ours) & \textbf{65.90} & \textbf{64.30} & \textbf{59.30} & \textbf{27.50} & \textbf{53.54}\\    
    \cline{2-7}
    & Y\textsubscript{base}$+$MDRC$+$Transformer & 45.80 & 52.40 & 48.00 & 16.60 & 50.37\\
    & Y\textsubscript{base}$+$MDRC$+$CBAM & \textbf{65.50} & 49.20 & 58.20 & 24.80 & 53.66\\
    & Y\textsubscript{base}$+$MDRC$+$ViT & 61.70 & 58.70 & 56.40 & 18.20 & 52.80\\
    & Y\textsubscript{base}$+$MDRC$+$SPP & 60.70 & 55.60 & 61.60 & 23.80 & 61.04\\
    & YOLO-DCAP (ours) & 60.00 & \textbf{73.70} & \textbf{65.40} & \textbf{28.10} & \textbf{69.74}\\
    \hline
    % \hline
    \multirow{12}{*}{BORE} %{\rotatebox{90}{BORE}}
    & Y\textsubscript{base} & 60.40 & 60.20 & 54.60 & 16.20 & 44.00 \\
    \cline{2-7}
    & Y\textsubscript{base}$+$MDRC (ours) & 61.30 & 58.70 & 56.10 & 18.60 & 47.40\\
    \cline{2-7}
    & Y\textsubscript{base}$+$Transformer & 72.10 & 60.00 & 57.20 & 18.50 & 58.20\\
    & Y\textsubscript{base}$+$CBAM & 65.70 & 59.30 & 57.00 & \textbf{22.20} & 55.22\\
    & Y\textsubscript{base}$+$ViT & 53.80 & 51.00 & 57.60 & 20.10 & 57.20\\
    & Y\textsubscript{base}$+$SPP & \textbf{71.00} & 55.60 & \textbf{58.10} & 21.30 & 58.45\\
    & Y\textsubscript{base}$+$AaSP (ours) & 52.50 & \textbf{60.60} & 57.70 & 22.10 & \textbf{60.50}\\    
    \cline{2-7}
    & Y\textsubscript{base}$+$MDRC$+$Transformer & \textbf{72.10} & 60.00 & 57.20 & 18.50 & 59.89\\
    & Y\textsubscript{base}$+$MDRC$+$CBAM & 62.80 & 57.10 & 54.60 & 19.80 & 56.81\\
    & Y\textsubscript{base}$+$MDRC$+$ViT & 64.50 & \textbf{67.20} & 59.80 & 20.70 & 59.10\\
    & Y\textsubscript{base}$+$MDRC$+$SPP & 68.40 & 65.10 & 60.80 & 23.30 & 62.30\\
    & YOLO-DCAP (ours) & 67.26 & 65.20 & \textbf{63.80} & \textbf{25.80} & \textbf{66.74}\\
    \hline
    % \hline
    \multirow{12}{*}{OCEAN EDDY} %{\rotatebox{90}{OCEAN EDDY}} 
    & Y\textsubscript{base} & 50.30 & 55.60 & 51.40 & 20.40 & 47.44\\
    \cline{2-7}
    & Y\textsubscript{base}$+$MDRC (ours) & 60.20 & 63.70 & 59.10 & 20.30 & 58.88\\
    \cline{2-7}
    & Y\textsubscript{base}$+$Transformer & 64.60 & 60.80 & 69.40 & 21.60  & 68.80\\
    & Y\textsubscript{base}$+$CBAM & 67.00 & 73.70 & 70.60 & 26.70 & 74.56\\
    & Y\textsubscript{base}$+$ViT & 59.20 & 70.00 & 71.20 & 29.60 & 71.86\\
    & Y\textsubscript{base}$+$SPP & \textbf{70.90} & \textbf{78.20} & 71.80 & 27.60 & 73.81\\
    & Y\textsubscript{base}$+$AaSP (ours) & 67.40 & 76.80 & \textbf{75.90} & \textbf{32.10} & \textbf{76.60}\\
    \cline{2-7}
    & Y\textsubscript{base}$+$MDRC$+$Transformer & 67.40 & 71.90 & 73.60 & 30.90 & 73.70 \\
    & Y\textsubscript{base}$+$MDRC$+$CBAM & 67.00 & 73.70 & 73.00 & 26.70 & 73.34 \\
    & Y\textsubscript{base}$+$MDRC$+$ViT & 63.30 & 75.40 & 75.00 & 32.50 & 80.47 \\
    & Y\textsubscript{base}$+$MDRC$+$SPP & 62.50 & \textbf{86.00} & 76.20 & 31.30 & 82.09 \\
    & YOLO-DCAP (ours) & \textbf{71.50} & 78.90 & \textbf{81.40} & \textbf{33.50} & \textbf{84.42} \\
    \hline
  \end{tabular}
  % \vspace{-2em}
\end{table*}

\section{Experiments, Results and Discussions}
\label{sec:exp}
% \vspace{-.25cm}
We evaluated our YOLO-DCAP model, which integrates MDRC and the attention-aided AaSP module, against YOLOv5 and state-of-the-art methods, including CBAM \cite{woo2018cbam}, Transformer \cite{vaswani2017attention}, SPP \cite{he2015spatial}, and ViT \cite{dosovitskiy2020image}, as shown in Table~\ref{tab:sota-comp}. The comparison spans three datasets, GW, Bore, and OE using precision, recall, mean average precision (mAP50 and mAP50-95), and intersection over union (IoU) metrics. Precision measures the accuracy of positive predictions, recall quantifies the detection of relevant objects, mAP50 and mAP50-95 is computed by averaging precision across IoU thresholds (0.5 and 0.5–0.95, respectively). IoU defined as $IoU = \frac{|A \cap B|}{|A \cup B|}$,
% \cite{IoU}, 
captures the overlap between predicted and actual regions. Notably, Table~\ref{tab:sota-comp} compares our proposed approaches, particularly the attention-based AaSP module, over SPP and other attention-based methods like CBAM, Transformer, and ViT.
% \vspace{-.75cm}
% \textbf{Data Preparation.} 

We utilized \texttt{LabelImg} tool
% \cite{LabelImg} 
to create bounding boxes around objects of interest, defined by their top-left $(x_1, y_1)$ and bottom-right $(x_2, y_2)$ coordinates. YOLO uses these labels to determine Intersection over Union (IoU) scores, evaluating the overlap between predicted and ground truth bounding boxes. Our dataset comprises 600 GW instances, 400 OE instances, and 870 Bore instances. For all models, we employed a 70:20:10 ratio for training, validation, and testing splits, respectively.

% \textbf{Evaluating Model Performances for All Datasets.}
\subsection{Evaluating Model Performances for All Datasets}

In terms of GW data, adding MDRC significantly improves mAP50 and IoU from the base YOLO model (denoted as Y\textsubscript{base}) by 6\% and nearly 8\%, respectively. The proposed AaSP improved over all other state-of-the-art and baseline methods by 13\% on average (by taking the difference of each from AaSP and averaging them). Among only attention-based methods, the IoU of our AaSP outperformed Transformer, CBAM, and ViT coupled with the Y\textsubscript{base} model. The IoU of SPP (coupled with Y\textsubscript{base}) is competitive with only 2\% less than our AaSP.
Comparing the proposed YOLO-DCAP (having MDRC and AaSP together) with all other state-of-the-art (e.g., Y\textsubscript{base}$+$MDRC$+$Transformer, etc.) and baseline methods, we see a significant enhancement of 9\% on average in the mAP50 score. For IoU, we observe a huge leap with the proposed YOLO-DCAP. Our approach outperformed SPP by nearly 9\% and surpassed all other methods, such as Transformer, CBAM, and ViT, by roughly 18\% on average. Considering the remaining metrics, like precision, recall, and mAP50-95, most of them showed improvement, though some fell slightly short. The mAP and IoU in the proposed YOLO-DCAP improved by around 23\% and 37\% respectively, over the base YOLO model.

Considering the Bore data, we observe a consistent trend with comparatively higher scores. However, adding MDRC to YOLO resulted in slight improvements, with mAP50 and IoU increasing by only 1.5\% and 3.4\%. While the AaSP module (coupled with Y\textsubscript{base}), showed competitive mAP scores compared to other baseline methods, it surpassed all approaches in IoU by 2\%. When evaluating the proposed YOLO-DCAP model, which combines MDRC and AaSP, against all other baseline methods (denoted as, Y\textsubscript{base}$+$MDRC$+$CBAM, etc.), we observe a noticeable improvement of over 6\% on average in the mAP50 score and approximately 7\% in the IoU score. For precision, recall, and mAP50-95, YOLO-DCAP demonstrated improvements in most cases, with some exceptions where recall and mAP50-95 reached their highest scores. Overall, YOLO-DCAP achieved remarkable gains, with mAP and IoU improving by 9.24\% and 22.74\% over the base YOLO (Y\textsubscript{base}) model.

Reflecting on the OE data, we observe an average improvement in performance across all models. The MDRC (Y\textsubscript{base}$+$MDRC) demonstrated enhanced scores compared to the Y\textsubscript{base} model. Considering the AaSP module alongside similar attention-based approaches and SPP, as well as the YOLO-DCAP model in comparison to all state-of-the-art methods (incorporating MDRC), the progression follows a trend of gradual improvement, with each method achieving highly competitive scores, indicating the MDRC is positively impacting the training process. Similar to the GW and Bore datasets, the AaSP module or YOLO-DCAP did not deviate significantly from other peer methods. However, our approaches, MDRC and AaSP individually, and their combination as YOLO-DCAP still outperformed all models in both mAP50 and IoU scores, with YOLO-DCAP achieving an impressive 84.42\% IoU.

\begin{table}[h]
  \centering
  % \small
  % \scriptsize
  % \footnotesize
  \caption{Mean and Standard Deviation comparison between baselines and the proposed YOLO-DCAP approaches.}
  \label{tab:mnstd}
  \begin{tabular}{|c|l|c|c|} % Left and right borders
    \hline
    % Dataset & Methods & Mean(mAP50\%) & Std. Dev.\\
    \textbf{Datasets} & \multicolumn{1}{c|}{\textbf{Methods}} & \textbf{mAP50(\%)} & \textbf{IoU(\%)}\\
    \hline
    \multirow{8}{*}{\shortstack{GRAVITY\\WAVE}}%{\rotatebox{90}{GRAVITY WAVE}} 
    & Y\textsubscript{base} & 36.38$\pm$5.42 & 28.87$\pm$5.78\\
    \cline{2-4}
    & Y\textsubscript{base}$+$MDRC (Ours) & 42.82$\pm$4.76 & 37.48$\pm$4.18\\
    & Y\textsubscript{base}$+$AaSP (Ours) & 55.80$\pm$4.59 & 50.28$\pm$3.02\\
    \cline{2-4}
    & Y\textsubscript{base}$+$MDRC$+$Transformer & 45.50$\pm$6.46 & 44.64$\pm$5.12\\
    & Y\textsubscript{base}$+$MDRC$+$CBAM & 53.60$\pm$4.63 & 50.18$\pm$3.11\\
    & Y\textsubscript{base}$+$MDRC$+$ViT & 52.20$\pm$4.46 & 48.76$\pm$4.14\\
    & Y\textsubscript{base}$+$MDRC$+$SPP & 56.50$\pm$3.15 & 58.72$\pm$3.18\\
    & YOLO-DCAP (Ours) & \textbf{63.18$\pm$1.58} & \textbf{68.18$\pm$1.82}\\
    \hline
    \multirow{8}{*}{BORE} %{\rotatebox{90}{BORE}} 
    & Y\textsubscript{base} & 51.54$\pm$3.71 & 40.56$\pm$3.94\\
    \cline{2-4}
    & Y\textsubscript{base}$+$MDRC (Ours) & 54.52$\pm$2.62 & 44.38$\pm$3.32\\
    & Y\textsubscript{base}$+$AaSP (Ours) & 55.70$\pm$2.22 & 58.14$\pm$2.06\\
    \cline{2-4}
    & Y\textsubscript{base}$+$MDRC$+$Transformer & 55.63$\pm$2.33 & 56.22$\pm$3.36\\
    & Y\textsubscript{base}$+$MDRC$+$CBAM & 53.14$\pm$1.24 & 53.79$\pm$2.67\\
    & Y\textsubscript{base}$+$MDRC$+$ViT & 56.36$\pm$2.64 & 56.82$\pm$2.84\\
    & Y\textsubscript{base}$+$MDRC$+$SPP & 58.18$\pm$2.40 & 59.88$\pm$3.32\\
    & YOLO-DCAP (Ours) & \textbf{62.34$\pm$0.91} & \textbf{65.28$\pm$1.02}\\
    \hline
    \multirow{8}{*}{\shortstack{OCEAN\\EDDY}} %{\rotatebox{90}{OCEAN EDDY}} 
    & Y\textsubscript{base} & 47.62$\pm$4.30 & 42.31$\pm$5.25\\
    \cline{2-4}
    & Y\textsubscript{base}$+$MDRC (Ours) & 55.54$\pm$3.89 & 54.86$\pm$4.20\\
    & Y\textsubscript{base}$+$AaSP (Ours) & 73.10$\pm$2.88 & 72.62$\pm$3.78\\
    \cline{2-4}
    & Y\textsubscript{base}$+$MDRC$+$Transformer & 69.60$\pm$3.92 & 68.24$\pm$4.28\\
    & Y\textsubscript{base}$+$MDRC$+$CBAM & 70.06$\pm$3.10 & 69.78$\pm$3.77\\
    & Y\textsubscript{base}$+$MDRC$+$ViT & 71.74$\pm$4.18 & 77.20$\pm$3.76\\
    & Y\textsubscript{base}$+$MDRC$+$SPP & 75.26$\pm$1.61 & 79.73$\pm$2.48\\
    & YOLO-DCAP (Ours) & \textbf{80.78$\pm$1.04} & \textbf{83.26$\pm$1.36}\\
    \hline
\end{tabular}
% \vspace{-2em}
\end{table}

% \subsection{Mean and standard deviation comparison}
We compared the mean and standard deviation across different configurations to assess consistency and reliability, as shown in Table~\ref{tab:mnstd}. The values were calculated based on 5 test runs per model, ensuring a thorough evaluation of performance stability. The YOLO-DCAP model achieved the highest mean mAP50 scores of 63.18\%, 62.34\%, and 80.78\% for GW, Bore, and OE, respectively, with the lowest standard deviations of $\pm$1.58, $\pm$0.91, and $\pm$1.04. For IoU scores, YOLO-DCAP similarly outperformed all models with 68.18\%, 65.28\%, and 83.26\% across the three datasets, with minimal deviations of $\pm$1.82, $\pm$1.02, and $\pm$1.36. Figure~\ref{fig:vis-res} demonstrates that YOLO-DCAP's satellite object localization closely aligns with labeled data, unlike other state-of-the-art models which often struggled with proper detection. Additionally, our task-oriented approaches (MDRC and AaSP) consistently outperformed the Y\textsubscript{base} model in both mean performance and standard deviation scores, highlighting YOLO-DCAP's superior accuracy and consistency for satellite object localization.

\begin{figure}%[ht!]
    \centering
    \includegraphics[width=.8\columnwidth]{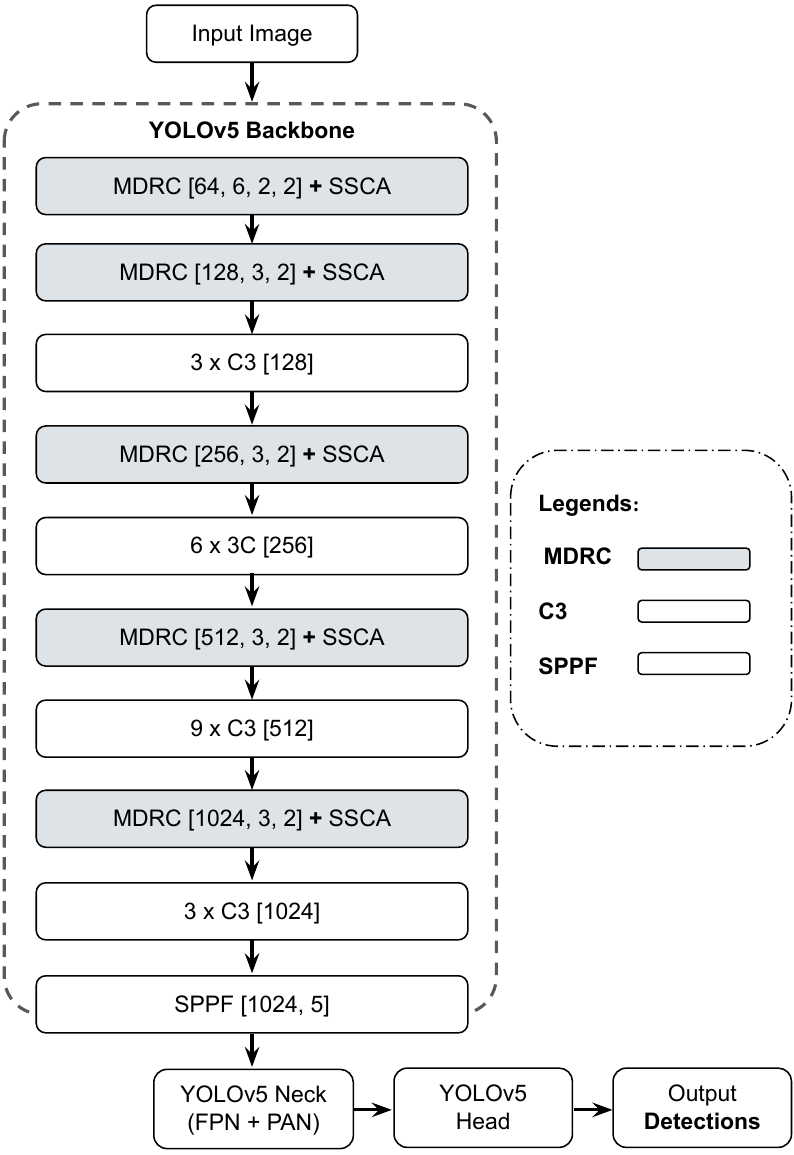}
    \caption{Modified YOLOv5 backbone with SSCA.}
    \label{fig:yolo-ssca}
% \vspace{-.5em}
\end{figure}

\subsection{Ablation study}
\begin{figure}[htb!]
    \centering
    \includegraphics[width=\columnwidth]{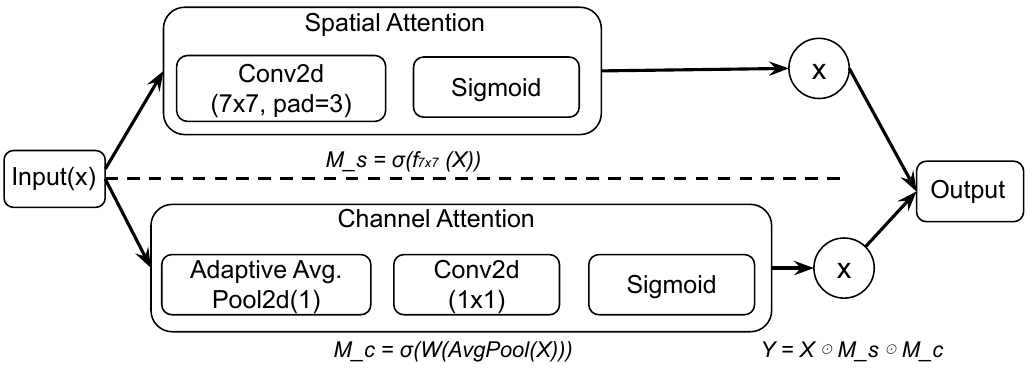}
    \caption{The Proposed SSCA architecture.}
    \label{fig:ssca}
% \vspace{-.5em}
\end{figure}

We explored an alternative approach to YOLO-DCAP using Simplified Spatial and Channel Attention (SSCA) implemented immediately after each Multi-scale Dilated Residual Convolutional (MDRC) block in the backbone. Figure~\ref{fig:yolo-ssca} represents the alternative YOLO backbone and Figure~\ref{fig:ssca} represents the SSCA architecture. This mechanism enhances important features while suppressing noise by focusing on relevant spatial areas and channels, thereby improving the model's ability to distinguish objects from background interference.

Inspired by CBAM \cite{woo2018cbam}, our SSCA approach streamlines feature extraction and addresses scale, shape, and pattern variability by simultaneously computing spatial and channel attention. CBAM applies channel and spatial attention sequentially with both average and max pooling, while Squeeze-and-Excitation focuses solely on channel relationships through global average pooling and fully connected layers (applied in AaSP). Spatial attention is computed by applying a \( 7 \times 7 \) convolution with padding on the input feature map, followed by a sigmoid activation, i.e., \( M_s = \sigma(f_{7 \times 7}(X)) \). Channel attention is derived from global average pooling followed by a \( 1 \times 1 \) convolution and sigmoid activation, i.e., \( M_c = \sigma(W(\text{AvgPool}(X))) \). The final output is computed as \( Y = X \odot M_s \odot M_c \), where \( \sigma(\cdot) \) denotes the sigmoid function and attention maps are broadcast and applied element-wise to \( X \).

This combined MDRC$+$SSCA mechanism effectively directs the network's focus to the most salient regions and channels, crucial for detecting objects in challenging satellite imagery. 
As shown in Table~\ref{tab:sota-comp2}, we compare SSCA with AaSP, both integrated with MDRC and built on top of the Y\textsubscript{base} backbone. While the AaSP-based configuration consistently outperforms SSCA across all three datasets and evaluation metrics including precision, recall, mAP50, mAP50-95, and IoU, the SSCA-based approach still delivers highly competitive performance. Notably, although SSCA does not surpass AaSP, it offers competitive performance and, in specific metrics or datasets, performs on par with or better than some existing attention-based methods such as Transformer. These comparative insights, highlighted in Table~\ref{tab:sota-comp}, suggest that a simplified attention mechanism like SSCA, when placed appropriately within the architecture, can still serve as an effective and efficient alternative for performance enhancement.

\begin{table*}[hb]
  \centering 
  \caption{Performance comparison of SSCA and AaSP (both combined with MDRC) across all datasets.}
  \label{tab:sota-comp2}
  \begin{tabular}{|c|l|c|c|c|c|c|}
    \hline
    \textbf{Datasets} & \multicolumn{1}{c|}{\textbf{Methods}} & \textbf{Precision} & \textbf{Recall} & \textbf{mAP50 (\%)} & \textbf{mAP50-95 (\%)} & \textbf{IoU(\%)}\\
    % \textbf{Datasets} & \textbf{Methods} & \textbf{Precision (\%)} & \textbf{Recall (\%)} & \textbf{mAP50 (\%)} & \textbf{mAP50-95 (\%)} & \textbf{IoU (\%)} \\
    \hline
    \multirow{2}{*}{GRAVITY WAVE} %{\rotatebox{90}{GRAVITY WAVE}} 
    % & Y\textsubscript{base}$+$MDRC$+$Transformer & 49.40 & 50.40 & 44.50 & 17.90 & 37.71 \\
    % & Y\textsubscript{base}$+$MDRC$+$ViT & 51.90 & 58.80 & 49.20 & 21.30 & 38.12 \\
    % & Y\textsubscript{base}$+$MDRC$+$CBAM & \textbf{59.50} & 59.10 & 52.80 & 23.80 & 44.48 \\
    % & Y\textsubscript{base}$+$AaSP (ours) & \textbf{65.90} & \textbf{64.30} & \textbf{59.30} & \textbf{27.50} & \textbf{53.54}\\
    & Y\textsubscript{base}$+$MDRC$+$SSCA & 58.80 & 66.70 & 55.30 & 26.60 & 48.74 \\
    & Y\textsubscript{base}$+$MDRC$+$AaSP & 60.00 & 73.70 & 65.40 & 28.10 & 69.74\\    
    \hline
    % \hline
    \multirow{2}{*}{BORE}%{\rotatebox{90}{BORE}}
    % & Y\textsubscript{base}$+$MDRC$+$Transformer & 44.90 & 46.80 & 40.20 & 16.10 & 33.28 \\
    % & Y\textsubscript{base}$+$MDRC$+$ViT & \textbf{55.90} & 62.60 & 51.40 & 23.80 & 41.82 \\
    % & Y\textsubscript{base}$+$MDRC$+$CBAM & 53.30 & 61.50 & 51.30 & \textbf{25.60} & 44.62 \\
    & Y\textsubscript{base}$+$MDRC$+$SSCA & 55.80 & 63.30 & 52.50 & 25.20 & 46.30 \\
    & Y\textsubscript{base}$+$MDRC$+$AaSP & 67.26 & 65.20 & 63.80 & 25.80 & 66.74\\
    \hline
    % \hline
    \multirow{2}{*}{OCEAN EDDY} %{\rotatebox{90}{OCEAN EDDY}}
    % & Y\textsubscript{base}$+$MDRC$+$Transformer & 55.60 & 52.80 & 58.20 & 33.10 & 57.44 \\
    % & Y\textsubscript{base}$+$MDRC$+$ViT & 59.60 & 53.70 & 61.70 & 36.40 & 61.21 \\
    % & Y\textsubscript{base}$+$MDRC$+$CBAM & \textbf{62.50} & 64.10 & 65.60 & 39.60 & 65.76 \\
    & Y\textsubscript{base}$+$MDRC$+$SSCA & 62.20 & 65.30 & 66.50 & 40.70 & 68.66 \\
    & Y\textsubscript{base}$+$MDRC$+$AaSP & 71.50 & 78.90 & 81.40 & 33.50 & 84.42 \\
    \hline
  \end{tabular}
\end{table*}

Table~\ref{tab:mnstd2} presents the mean and standard deviation analysis for mAP50 and IoU metrics across three datasets, comparing the SSCA and AaSP modules when integrated with MDRC into the Y\textsubscript{base} backbone. While the AaSP-based configuration consistently delivers the best performance with low variability, the SSCA-based setup remains competitive, achieving mAP50 scores of 54.56\%, 50.86\%, and 65.08\%, and IoU values of 46.78\%, 44.66\%, and 67.84\% for Gravity Wave, BORE, and Ocean Eddy, respectively. These results highlight SSCA’s stability, with relatively low variance across datasets. Although SSCA does not outperform ViT or CBAM in few cases, it consistently delivers better performance than Transformer on the Gravity Wave dataset and demonstrates the potential of a simplified attention mechanism when placed effectively within the architecture. This suggests that SSCA can serve as an efficient and lightweight alternative where model simplicity and consistency are desired.

We further evaluated the impact of multi-scale dilation ($d=2,$ $3$) on different layers such as replacing either convolutional layers or cross convolutional (C3) layers of the YOLO backbone to observe the model's performance across all datasets. Table~\ref{tab:dilation-effects} demonstrates that implementing MDRC in convolutional layers significantly outperforms the baseline models for all datasets. However, MDRC in C3 layers showed notably reduced performance across all metrics. It is our assumption that the C3 layers, with their existing cross-stage connections and bottleneck structures, may create conflicts when combined with dilated convolutions, potentially disrupting the feature extraction process. 

\begin{table}
  \centering
  \caption{Mean and Standard Deviation Comparison between Y\textsubscript{base}$+$MDRC$+$AaSP (proposed as, YOLO-DCAP) and Y\textsubscript{base}$+$MDRC$+$SSCA approaches.}
  \label{tab:mnstd2}
  \begin{tabular}{|c|l|c|c|} % Left and right borders
    \hline
    \textbf{Datasets} & \multicolumn{1}{c|}{\textbf{Methods}} & \textbf{mAP50(\%)} & \textbf{IoU(\%)}\\
    \hline
    \multirow{2}{*}{\shortstack{GW}} %{\rotatebox{90}{GRAVITY WAVE}} 
    % & Y\textsubscript{base}$+$MDRC$+$TR. & 41.60$\pm$4.80 & 35.71$\pm$3.50 \\
    % & Y\textsubscript{base}$+$MDRC$+$ViT & 46.10$\pm$3.30 & 36.42$\pm$2.60 \\
    % & Y\textsubscript{base}$+$MDRC$+$CBAM & 50.90$\pm$2.40 & 42.68$\pm$2.20 \\
    & Y\textsubscript{base}$+$MDRC$+$SSCA & 54.56$\pm$1.20 & 46.78$\pm$1.90 \\
    & Y\textsubscript{base}$+$MDRC$+$AaSP & 63.18$\pm$1.58 & 68.18$\pm$1.82\\
    \hline
    \multirow{2}{*}{BORE} %{\rotatebox{90}{BORE}} 
    % & Y\textsubscript{base}$+$MDRC$+$TR. & 35.20$\pm$4.92 & 27.50$\pm$5.48 \\
    % & Y\textsubscript{base}$+$MDRC$+$ViT & 46.76$\pm$4.42 & 37.58$\pm$4.30 \\
    % & Y\textsubscript{base}$+$MDRC$+$CBAM & 47.12$\pm$3.88 & 40.22$\pm$3.85 \\
    & Y\textsubscript{base}$+$MDRC$+$SSCA & 50.86$\pm$1.56 & 44.66$\pm$1.72 \\
    & Y\textsubscript{base}$+$MDRC$+$AaSP & 62.34$\pm$0.91 & 65.28$\pm$1.02\\
    \hline
    \multirow{2}{*}{\shortstack{OE}} %{\rotatebox{90}{OCEAN EDDY}}
    % & Y\textsubscript{base}$+$MDRC$+$TR. & 55.26$\pm$3.77 & 55.42$\pm$4.09 \\
    % & Y\textsubscript{base}$+$MDRC$+$ViT & 59.47$\pm$2.61 & 57.78$\pm$3.21 \\
    % & Y\textsubscript{base}$+$MDRC$+$CBAM & 64.10$\pm$2.48 & 63.83$\pm$2.32 \\
    & Y\textsubscript{base}$+$MDRC$+$SSCA & 65.08$\pm$1.16 & 67.84$\pm$1.48 \\
    & Y\textsubscript{base}$+$MDRC$+$AaSP & 80.78$\pm$1.04 & 83.26$\pm$1.36\\

    \hline
\end{tabular}
\end{table}

Additionally, we evaluated the effects of a single dilation rate ($d=2$). Table~\ref{tab:dialation2} evaluates the effect of using a single dilation rate compared to the multi-scale approach in our proposed models. The results indicate that using a single dilation rate with MDRC yields lower performance than the multi-scale dilation approach reported in our primary results. However, even with a single dilation rate, adding attention mechanisms (either in SSCA or AaSP) consistently improves performance across all datasets. Notably, Y\textsubscript{base}$+$MDRC$+$AaSP outperforms Y\textsubscript{base}$+$MDRC$+$SSCA in both mAP50 and IoU metrics across all datasets.

\begin{table}
  \centering
    \caption{Effects of multi-scale dilation impact on different layers of YOLO backbone across all datasets.}
  \label{tab:dilation-effects}
  \begin{tabular}{|c|l|c|c|} 
    \hline
    \textbf{Datasets} & \multicolumn{1}{c|}{\textbf{Dilation Effects}} & \textbf{mAP50(\%)} & \textbf{IoU(\%)}\\
    \hline
    \multirow{3}{*}{GW}%{\shortstack{GRAVITY\\WAVE}}
    & Y\textsubscript{base} (No dilation) & 41.80 & 32.76 \\
    & MDRC in C3 layers ($d=2,3$) & 34.20 & 27.75 \\
    & MDRC in Conv layers ($d=2,3$) & \textbf{47.80} & \textbf{40.62} \\
    \hline
    \multirow{3}{*}{BORE} 
    & Y\textsubscript{base} (No dilation) & 54.60 & 44.00 \\
    & MDRC in C3 layers ($d=2,3$) & 41.40 & 31.00 \\
    & MDRC in Conv layers ($d=2,3$) & \textbf{56.10} & \textbf{47.40} \\
    \hline
    \multirow{3}{*}{OE}%{\shortstack{OCEAN\\EDDY}}
    & Y\textsubscript{base} (No dilation) & 51.40 & 47.44 \\
    & MDRC in C3 layers ($d=2,3$) & 42.20 & 38.94 \\
    & MDRC in Conv layers ($d=2,3$) & \textbf{59.10} & \textbf{58.88} \\
    \hline
  \end{tabular}
\end{table}

\begin{table}[h!]
  \centering
  % \scriptsize
% \begin{minipage}{0.4\linewidth}\centering
  \caption{Comparing proposed approaches (MDRC with SSCA, AaSP) with YOLO using a single dilation rate
  ($d=2$).}
  % shows comparatively lesser performance improvement, however, it is observed that adding an attention mechanism still improves the performance.}
  \label{tab:dialation2}
  \begin{tabular}{|c|l|c|c|} % Left and right borders
    % \hline
    % Datasets & Approach & mAP50(\%) & IoU(\%)\\
    \hline
    \textbf{Datasets} & \multicolumn{1}{c|}{\textbf{Methods}} & \textbf{mAP50(\%)} & \textbf{IoU(\%)}\\
    \hline
    \multirow{3}{*}{GW}%{\shortstack{GRAVITY\\WAVE}}
    & Y\textsubscript{base}$+$MDRC & 42.20 & 33.91 \\
    & Y\textsubscript{base}$+$MDRC$+$SSCA & 51.50 & 48.28 \\
    & Y\textsubscript{base}$+$MDRC$+$AaSP & \textbf{55.80} & \textbf{53.55} \\
    \hline
    \multirow{3}{*}{BORE} 
    & Y\textsubscript{base}$+$MDRC & 54.50 & 45.88 \\
    & Y\textsubscript{base}$+$MDRC$+$SSCA & 55.46 & 49.26 \\
    & Y\textsubscript{base}$+$MDRC$+$AaSP & \textbf{57.40} & \textbf{56.68} \\
    \hline
    \multirow{3}{*}{OE}%{\shortstack{OCEAN\\EDDY}}
    & Y\textsubscript{base}$+$MDRC & 53.80 & 52.64 \\
    & Y\textsubscript{base}$+$MDRC$+$SSCA & 63.44 & 60.66 \\
    & Y\textsubscript{base}$+$MDRC$+$AaSP & \textbf{72.30} & \textbf{71.38} \\
    \hline
  \end{tabular}
\end{table}

\section{Discussions and Conclusions}
\label{sec:disc-and-conc}
\subsection{Discussions}
This study addresses critical challenges inherent in satellite object detection, including significant variability in object scales, diverse shapes, intricate patterns, and complexities arising from occlusion and overlapping objects. Experimental evaluations confirm that the proposed YOLO-DCAP, integrating Multi-scale Dilated Residual Convolutional (MDRC) blocks and Attention-aided Spatial Pooling (AaSP), consistently outperforms the baseline YOLO model and several leading attention-based architectures (Transformer, CBAM, ViT), including SPP. YOLO-DCAP achieves substantial average improvements of approximately 20.95\% in mAP50 and 32.23\% in IoU over the base model, and approximately 7.35\% and 9.84\%, respectively, over other state-of-the-art methods across all the datasets in this study.

Our exploration of dilation rates underscores the importance of multi-scale feature extraction. Employing multiple dilation rates ($d=2,$ $3$) significantly enhances feature representation, outperforming single dilation configurations. Although single dilation still benefits from attention mechanisms (either AaSP or SSCA), multi-scale dilation strategies comprehensively address satellite imagery's inherent complexity.

The AaSP module notably extends and improves traditional Spatial Pyramid Pooling (SPP) by introducing attention-guided spatial pooling. AaSP selectively emphasizes critical spatial information, enhancing robustness against occlusion and overlaps, which are common challenges in satellite imagery. Our experiments demonstrate AaSP’s clear superiority over standard SPP and other established attention mechanisms. Additionally, the Simplified Spatial and Channel Attention (SSCA) mechanism offers a lightweight yet competitive alternative, particularly surpassing Transformer-based methods on specific datasets. The ablation studies further reveal potential conflicts between dilated convolutions and the existing bottleneck structures of C3 layers, reinforcing the strategic placement of MDRC in standard convolutional layers for effective feature extraction.

Overall, the consistent performance and reliability of YOLO-DCAP across diverse satellite phenomena (Gravity Waves, Bores, and Ocean Eddies) highlight its suitability for operational satellite monitoring systems. 

In the future, we would like to focus on extending our approach to a wider range of atmospheric phenomena, including Meso-scale Hurricane \cite{hasan2025comparison} and cloud properties \cite{tushar2024cloudunet}. We also plan to explore integration with advanced architectures such as latest YOLO architectures (v11, v12) or Detection Transformers (DETR) which we acknowledge as the current limitation of this work.

\subsection{Conclusion}
\label{subsection:conc}
We addressed the challenges of object localization in satellite imagery across three diverse phenomena: Gravity Waves, mesospheric Bores, and Ocean Eddies in this work. Our proposed novel approach, YOLO-DCAP, integrates Multi-scale Dilated Residual Convolution (MDRC) and Attention-aided Spatial Pooling (AaSP) to enhance feature extraction and refinement. Experimental results demonstrate that YOLO-DCAP consistently outperforms both baseline YOLO and state-of-the-art attention-based approaches, with substantial improvements in IoU scores indicating superior localization precision. The model's success across diverse phenomena and low standard deviations in performance metrics highlight its potential as a robust solution for satellite monitoring systems. This work advances satellite object detection for climate-related analysis and Earth informatics, establishing a new standard in computer vision research. With its strong performance, YOLO-DCAP shows significant potential for extension to additional atmospheric phenomena and emerging detection architectures.

\bibliographystyle{ieeetr}
\bibliography{egbib}

\end{document}